\documentclass[12pt]{article}
\usepackage{amssymb,graphicx}
\title{Deepest Neural Networks}
\author{Ra\'ul Rojas\\ Dahlem Center for Machine Learning and Robotics\\Freie Universit\"at Berlin}
\date{July 2017}
\begin{document}
\maketitle

\abstract{This paper shows that a long chain of perceptrons  (that is, a multilayer perceptron, or MLP, with many hidden layers of width one) can be a universal classifier. The classification procedure is not necessarily computationally efficient, but the technique throws some light on the kind of computations possible with narrow and deep MLPs.}

\section{Motivation}

In some classification problems we start with data from an input space $A$ of dimension $n$, where each point $(x_1,x_2,\ldots,x_n)$ in $A$ belongs either to a positive class $P$ (for example digits), or a negative class $N$ (for example nondigits). Sometimes we can separate both classes using a linear cut. In order to do that, we look for weights $(w_1,w_2,\ldots,w_n)$ and a threshold $\theta$, such that for all points in the positive class the inequality
$$
\sum_1^n{w_ix_i}>\theta
$$
holds, while for all points in the negative class the inequality does not hold. A small machine implementing this computation is called a ``perceptron''. We can think of a perceptron as a small unit where each component $x_i$ flows through a conduction channel with weight $w_i$. All incoming weighted information is added and compared to the threshold $\theta$ in the body of the unit. The ouput is binary: a 1 for points in the positive class, 0 for points in the negative class.

\begin{figure}[h]
\centerline{\includegraphics[width=9cm]{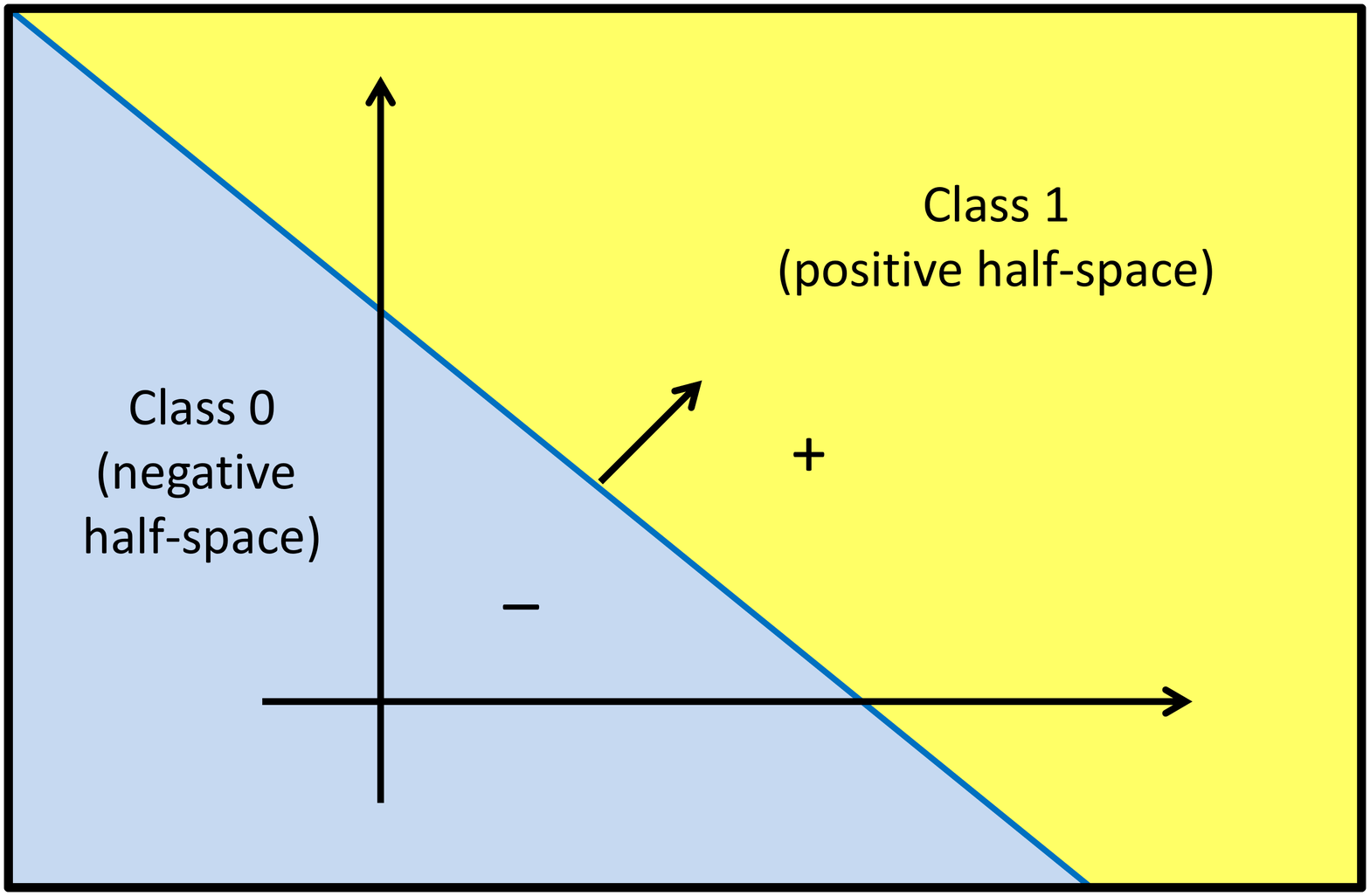}}
\caption{A perceptron divides the input space into two halves.\label{fig1}}
\end{figure}

Fig.~\ref{fig1} shows a linear cut in input space, such as those produced by a perceptron. The cut divides the space into two halves, one for class P and another for the class N. The cut is produced in Fig.~\ref{fig1} by a line. In general, the cut is produced by a hypersurface of dimension $n-1$. We adopt the convention here that the small vector drawn normal to the hypersurface points in the direction of the positive class.

\begin{figure}[h]
\centerline{\includegraphics[width=8cm]{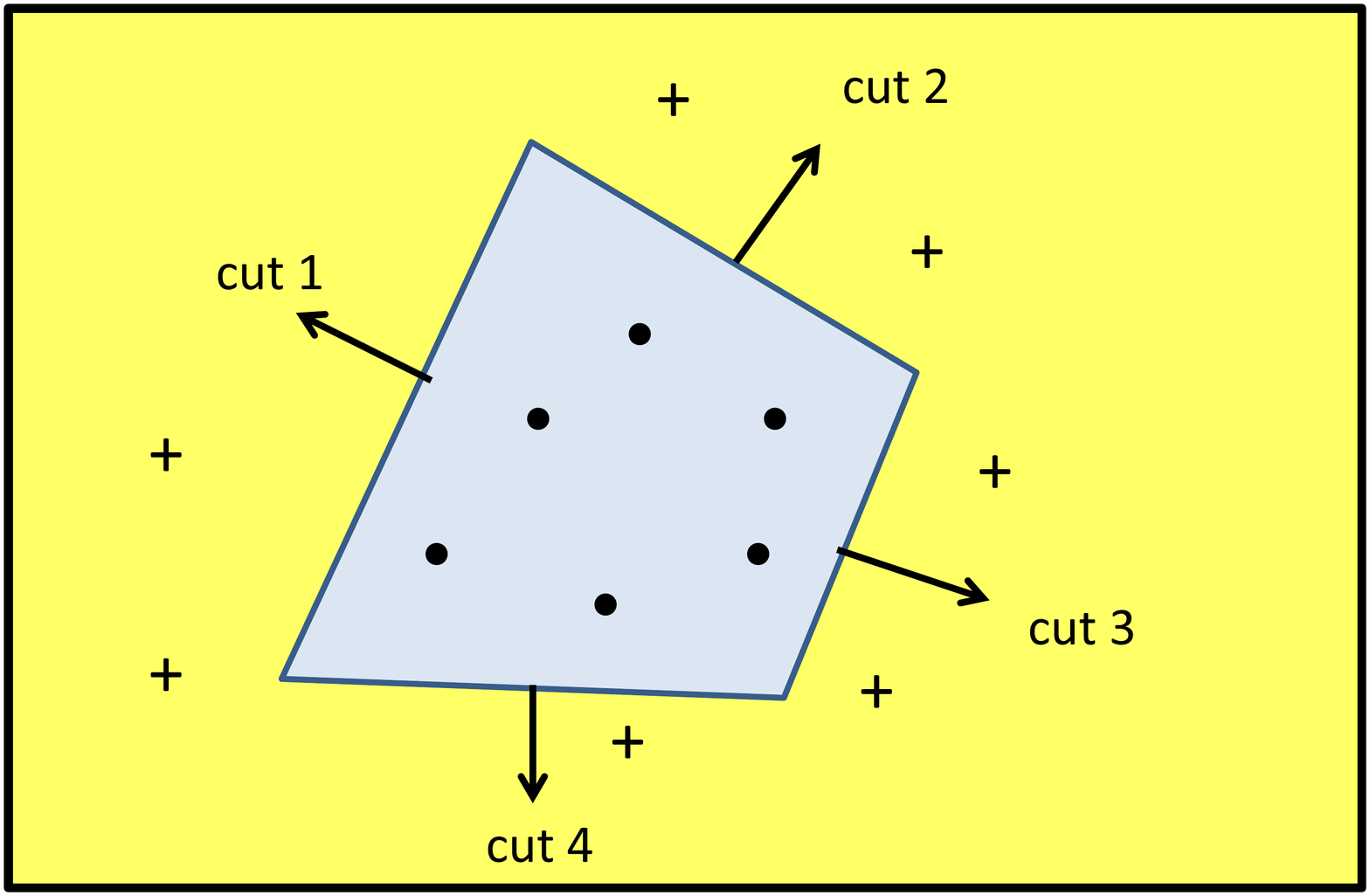}}
\caption{A convex polytope is limited by linear cuts. The polytope encloses the class represented by black circles, separating it from the crosses.\label{fig2}}
\end{figure}

Fig.~\ref{fig2} shows an example, where we want to separate a set of small circles from the crosses. A linear cut would not be sufficient, but in this case we can enclose the class of small circles using a convex polytope delimited by four linear cuts. In this case we could find four perceptrons whose weights are such that the intersection of the four negative half-spaces of the four cuts represents the region inside  the polytope (and therefore the class of small circles), while its complement represents the class of crosses. In this example we would have to test the output of four perceptrons, one for each linear cut. In MLPs this is usually done in parallel in a ``hidden layer''.

For simplicity, we will assume that the input to a perceptron is given by  vectors $(x_1,x_2,\ldots,x_n,1)$, where the $x_i$ represent the data and the additional 1 is an extra constant. The weights  of the perceptron are then represented by $(w_1,w_2,\ldots,w_n,-\theta)$ and the  inequality to be tested is 
$$
\sum_1^{n+1}{w_ix_i}>0
$$
which for  $x_{n+1}=1$ and $w_{n+1}=-\theta$ is equivalent to the inequality given before. With this simplification the threshold of the perceptrons we use is always zero, unless we specify otherwise.

\section{Testing convex regions}

\begin{figure}[htb]
\centerline{\includegraphics[width=12cm]{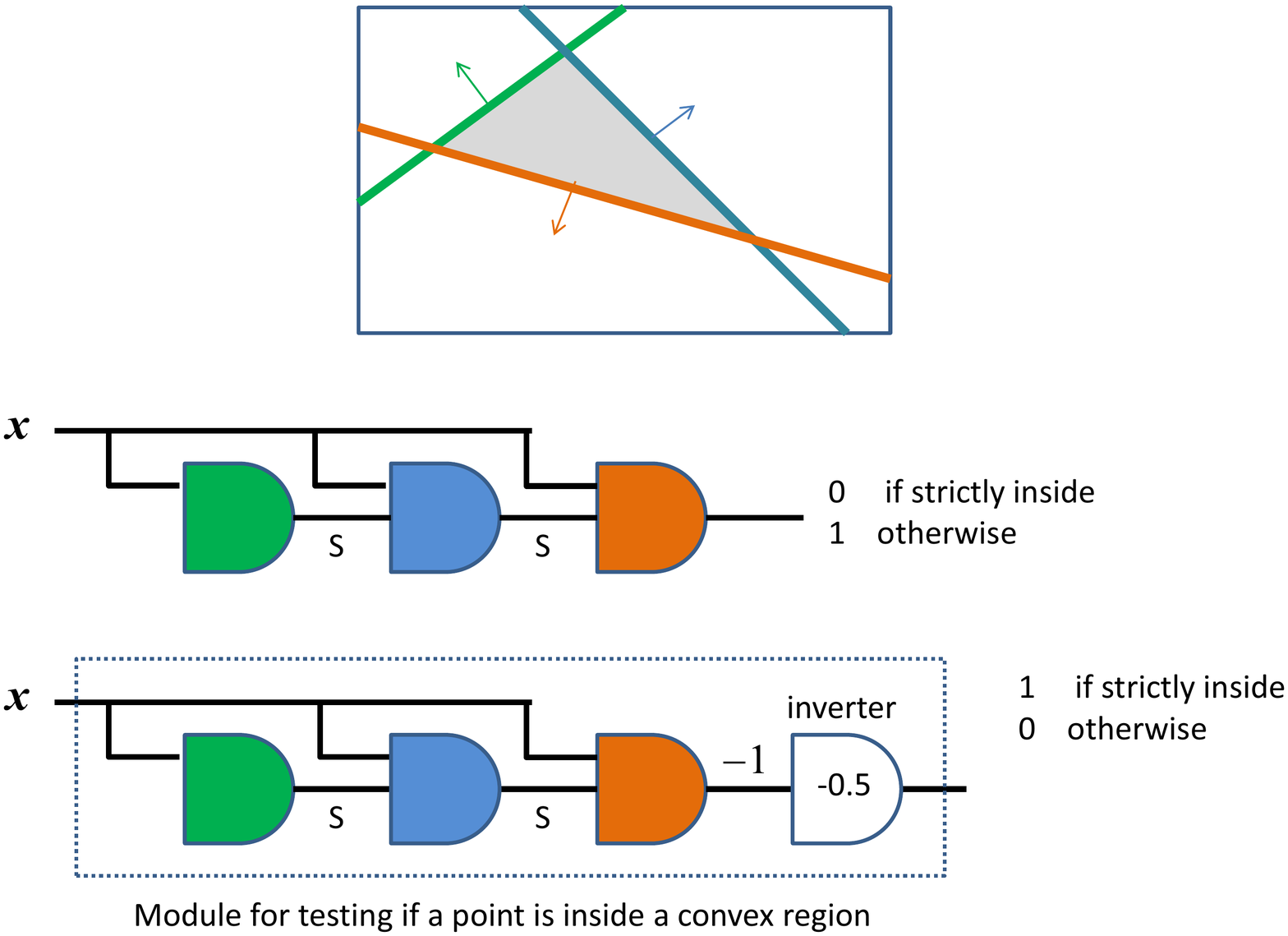}}
\caption{A chain of perceptrons can test if a point is inside a convex polytope.\label{fig3}}
\end{figure}

Fig.~\ref{fig3} shows, in three steps, how to test if a point is inside a convex polytope or not, using a chain of perceptrons. Assume that the region to be tested is delimited by the three lines shown in the upper part of Fig.~\ref{fig3}. We need three linear cuts represented by the perceptrons arranged in a chain in the middle of Fig.~\ref{fig3}. The input to the first perceptron is the $(n+1)$-dimensional vector $x$. The graphical convention here is that the input channel to each perceptron contains the necessary weights $(w_1,\ldots,w_{n+1})$, which are multiplied pairwise with the components of the input vector, and then the threshold operation (with threshold 0) is computed in the body of the perceptron.

We are assuming here that the data set comes from a bounded region of space. In images, for example, there is a maximum length of the input vectors (for example, all pixel values in an image are 1,  where the pixel values run from 0 to 1). We can always multiply the weights of a perceptron by a constant $\alpha$ so that  
$$
\|{\sum_1^{n+1}\alpha{w_ix_i}}\|<1
$$
holds, that is, the weighted input is not large in absolute value. Such scaling of the weights does not change the position of the linear cut in input space. Therefore, scaling the weights of all perceptrons, so that the inner products of data and weights are small, in absolute value, does not affect the shape nor location of the convex polytope.

The first perceptron in the chain in the middle of Fig.~\ref{fig3} gets as input only the data vector. If the data is outside the polytope, as tested by the first linear cut, this unit fires a 1, otherwise it outputs 0 and we should continue testing. But if the output of the first perceptron is 1, we do not have to continue testing the other two linear cuts (since we now know that the data point cannot possibly belong to the convex region being examined). Nevertheless, the chain of perceptrons will continue testing, but the second perceptron receives $x$ as input and also the output of the first perceptron, multiplied by the weight $S$. This weight $S$ is just a big positive number, much larger than $|\sum_1^{n+1}\alpha{w_ix_i}|$. This means that the sum of the weighted input and $S$ will always be larger than zero. The second perceptron must fire and its output 1 is again multiplied by $S$, making the third perceptron fire, and so on. The idea is that whenever one of the perceptrons in the chain fires, all other perceptrons along the chain will fire too, regardless of the input $x$.

If the first perceptron does not fire, we then test if the data point can be inside the polytope as defined by the second cut. Notice that now the weigth $S$ is being multiplied with 0 and does not interfere with the linear cut computation. In case that all perceptrons produce the output 0, that means that the data point is in the negative half-space of every single cut, that is, inside the convex polytope.

After this chain of computations, we put an inverter at the end, so that the complete chain of perceptrons produces a 1 for data points inside the convex region and 0 otherwise. Notice that the inverter uses a negative input weight and the threshold $-0.5$. Only when the output of the chain of three perceptrons is 0, is the input to the inverter greater than the threshold and this last inverter perceptron fires.

The reader should now be convinced that as long as the two classes to be separated are such that one of them can be fully encapsulated in a convex polytope, we can arrange a chain of perceptrons to provide the necessary classification.

\section{Nested polytopes}

In the general case, where the data does not allow a simple separation using one complex polytope, we can arrange for several polytopes to do the trick. The usual approach in MLPs is to corral one class, for example the positive class, using several disjoint convex polytopes. We can then use two, three or more polytopes, enclose one class with them, and then test every polytope in parallel. This is what MLPs usually do with their interplay between hidden and output layer. Here we need another approach since all hidden layers are of width one (that is, a single perceptron).

\begin{figure}[h]
\centerline{\includegraphics[width=9cm]{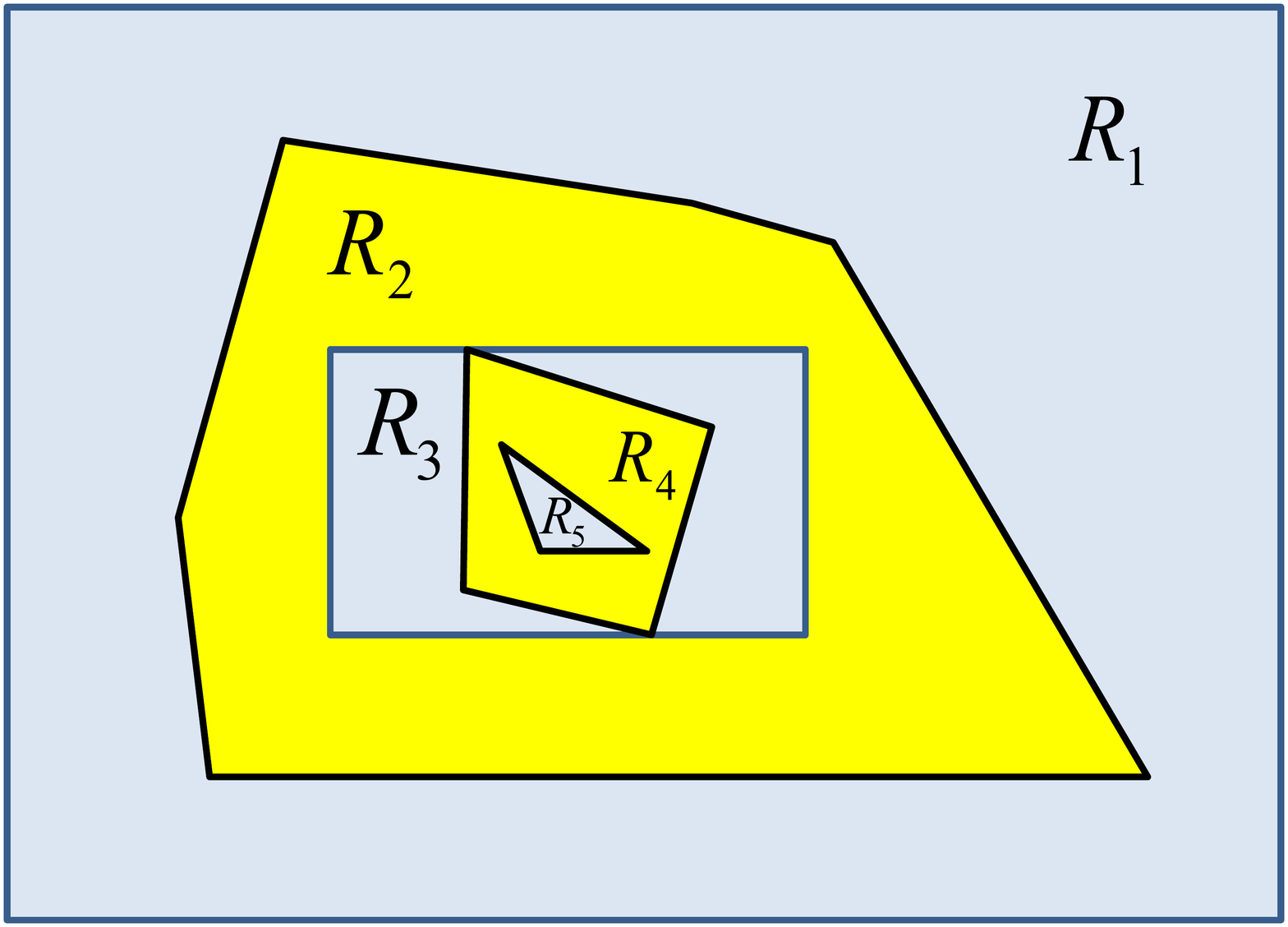}}
\caption{Two classes can be separated by nested convex regions.\label{fig4}}
\end{figure}

Fig.~\ref{fig4} shows what we can do. Assume that we have two classes, blue and yellow data points. Without loss of generality, assume that R$_1$ is the convex hull of the blue points (and encloses also all yellow points). We now find a region $R_2$ which is the convex hull of the yellow points inside $R_1$. We then find the region $R_3$ which is the convex hull of all blue points inside $R_2$. We proceed to find the region $R_4$ which is the convex hull of all yellow data points inside $R_3$, and so on, as shown in the figure. This succession of nested hulls provides  a way of classifying new unseen points from the data set. Here, I assume that there are no points which are both blue and yellow. If they exist, they are deleted from the training set, or we select one of them randomly.

Notice that we are not trying to be efficient here. Many examples can be given of nested convex regions where we advance too slowly and a new nested region is arbitrarily close to the previous one. But this shall not deter us, since we are providing a proof in principle, and the described procedure terminates.

\begin{figure}[h]
\centerline{\includegraphics[width=11cm]{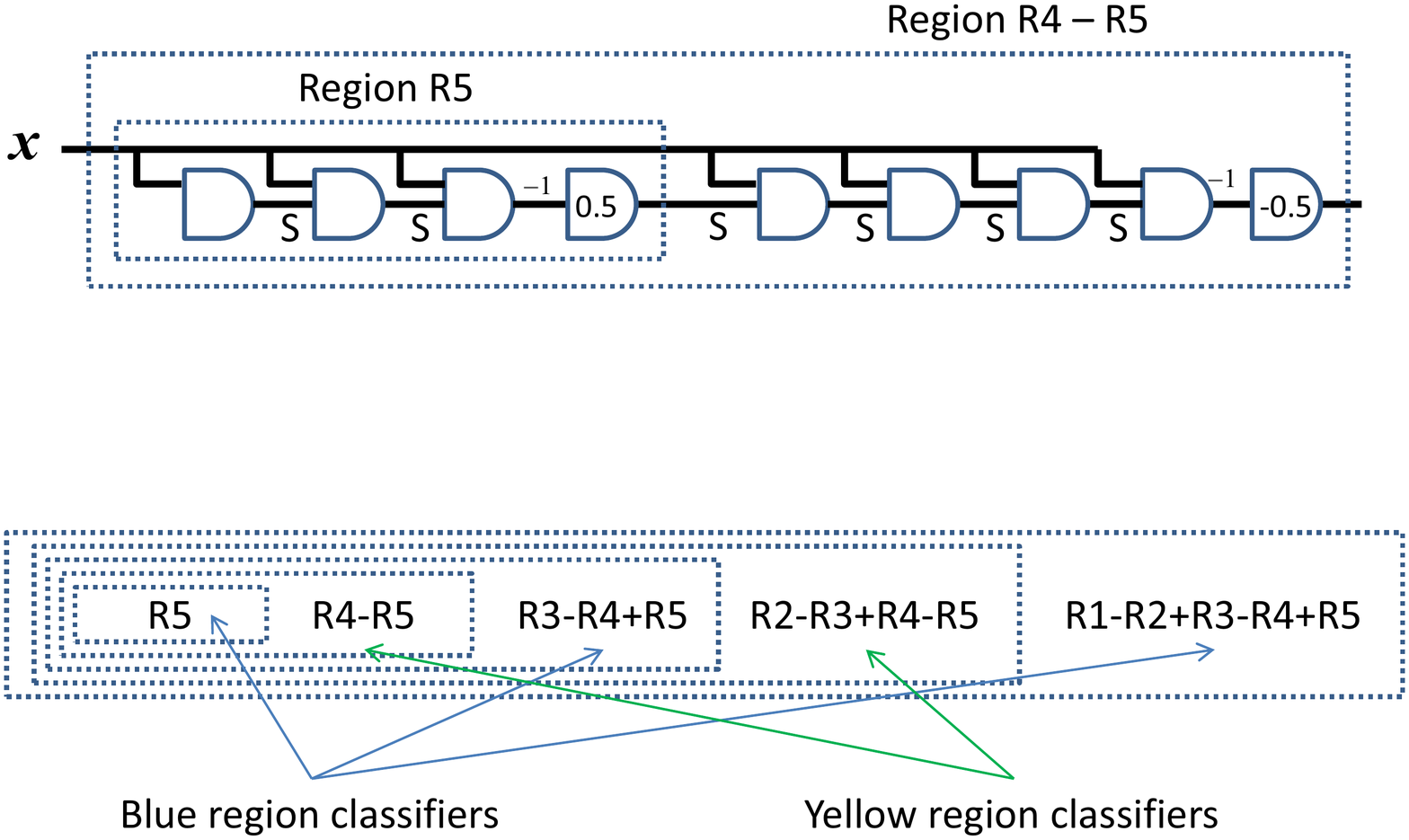}}
\caption{Deep perceptron chains for testing region membership.\label{fig5}}
\end{figure}

Given a succession of nested regions for the two classes blue and yellow, we can now find weights for a chain of perceptrons which can separate both of them. This is shown in Fig.~\ref{fig5}.

What we have to do is first classify region R5, the innermost region produced by the nesting approach described above. This can be done with a chain of perceptrons, as shown in the previous section. This chain is now a module for the recognition of points inside $R_5$. Now we put this module at the beginning of a new chain for recognizing points inside region $R_4$. If a point is inside $R_5$ it makes all perceptrons later in the chain for $R_4$ to fire. That is, now the perceptrons in the chain for region $R_4$ fire if the point $x$ is inside $R_5$, or outside $R_4$ (as tested by the four following perceptrons in the chain). An inverter at the end provides us with a module for the recognition of data points inside $R_4-R_5$.

We can continue with this approach and use the module for $R_4-R_5$ in order to produce a module for $R_3-(R_4-R_5)$, that is, for the region $R_3-R_4+R_5)$, and so on. At the end, using this modular approach, we can obtain a module for the region $R_1-R_2+R_3-R_4+R_5$, which can separate the blue from the yellow points.

\section{Discussion}

There is a well-known result which states that an MLP with a wide hidden layer and a single output
is a ``universal classifier''\cite{hastie}. Usually, smooth and continous output units are used for the proof (since we 
want to approximate smooth functions). A related result is that perceptrons, such as the ones described here, can separate any two classes using a hidden layer for defining many linear cuts, a second hidden layer for defining convex polytopes, and an ouput unit which fires if a point is inside any of the disjoint polytopes.
\begin{figure}[htb]
\centerline{\includegraphics[width=9cm]{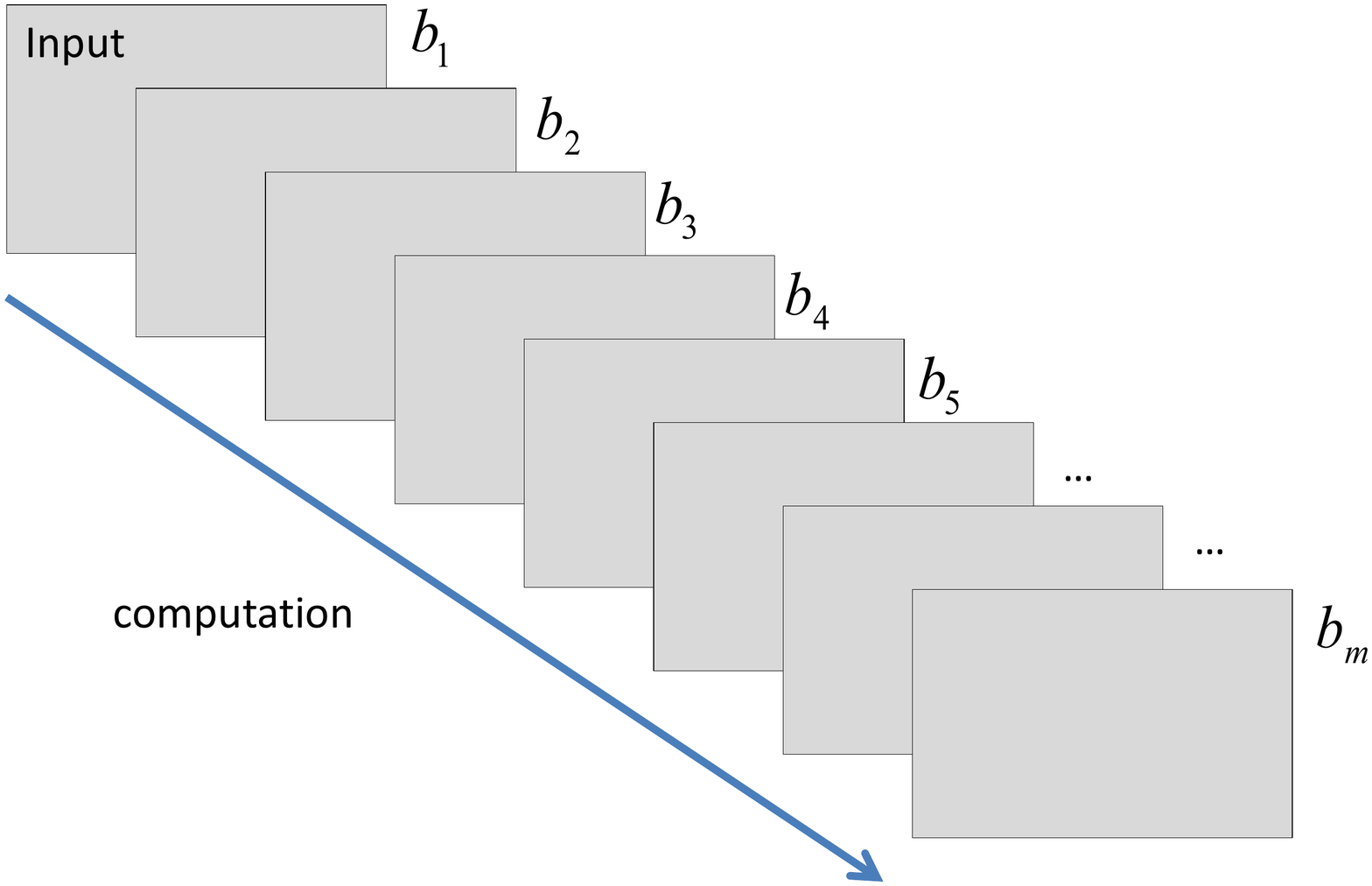}}
\caption{Deep perceptron chains for testing region membership.\label{fig6}}
\end{figure}
The result proved here is the ``dual'' of that well-known fact, in the sense that we can trade depth
for width. Instead of building a shallow network with wide hidden layers, we build a deep network with minimal width. Notice that the input vector is passed from layer to layer. This is called a ``shortcut'' in the neural networks literature and some deep neural networks also pass the original input from layer to layer.

The result might seem trivial but it is more intriguing if you think of it by using Fig.~\ref{fig6} as a kind of cartoon of a deep neural network computation. The idea is that you have an image that you want to classify. You pass the image from stage to stage of the deep computation but you can only pass one new bit of computation at every stage ($b_1, b_2, \ldots, b_m$). Can you classify correctly the image as a digit or non-digit? If the input space can be divided using nested polytopes, you can, as has been shown. Of course, the bit you pass is the one telling you if you can already discard the image as belonging to the positive class. If you can never discard the image until the computation ends, it then belongs to the positive class.

This result was presented in 2003 at a conference \cite{rojas}. I rewrote the paper to make it easier to read now that deep neural networks are so popular. Of course, the architecture described here does not progress to higher and higher levels of abstraction in each new layer, but it  nevertheless represents an interesting ``folk theorem''.

\end{document}